\documentclass{article}

\usepackage{microtype}
\usepackage{graphicx}
\usepackage{subfigure}
\usepackage{booktabs} 

\usepackage{hyperref}

\usepackage[accepted]{icml2021}
\usepackage{alp}

\icmltitlerunning{Stochastic Iterative Graph Matching}

\begin{document}

\twocolumn[
\icmltitle{Stochastic Iterative Graph Matching}




\begin{icmlauthorlist}
\icmlauthor{Linfeng Liu}{tuftscs}
\icmlauthor{Michael C. Hughes}{tuftscs}
\icmlauthor{Soha Hassoun}{tuftscs,tuftscb}
\icmlauthor{Li-Ping Liu}{tuftscs}
\end{icmlauthorlist}

\icmlaffiliation{tuftscs}{Department of Computer Science, Tufts University, MA, USA} 
\icmlaffiliation{tuftscb}{Department of Chemical and Biological Engineering, Tufts University, MA, USA}

\icmlcorrespondingauthor{Linfeng Liu}{linfeng.liu@tufts.edu}

\icmlkeywords{Machine Learning, ICML}

\vskip 0.3in
]



\printAffiliationsAndNotice{}  

\begin{abstract}
Recent works apply Graph Neural Networks (GNNs) to graph matching tasks and show promising results. Considering that model outputs are complex matchings, we devise several techniques to improve the learning of GNNs and obtain a new model, Stochastic Iterative Graph MAtching (SIGMA). Our model predicts a distribution of matchings, instead of a single matching, for a graph pair so the model can explore several probable matchings.
We further introduce a novel multi-step matching procedure, which learns how to refine a graph pair's matching results incrementally. The model also includes dummy nodes so that the model does not have to find matchings for nodes without correspondence. We fit this model to data via scalable stochastic optimization. We conduct extensive experiments across synthetic graph datasets as well as biochemistry and computer vision applications. Across all tasks, our results show that SIGMA can produce significantly improved graph matching results compared to state-of-the-art models. Ablation studies verify that each of our components (stochastic training, iterative matching, and dummy nodes) offers noticeable improvement.
\end{abstract}

\section{Introduction}

Graph matching  \citep{livi2013graph,yan2016short,sun2020survey} aims to find node correspondence among two or more graphs. It has a wide range of applications such as computer vision \citep{sun2020survey}, computational biology \citep{saraph2014magna}, and biochemistry \citep{kotera2004computational}. 
Given that many practical graphs cannot be perfectly matched, graph matching often maximizes some matching objective, such as the total number of matched edges \citep{yan2020learning}. 

Learning-based graph matching \citep{caetano2009learning,zanfir2018deep,yu2019learning,wang2019learning,fey2018splinecnn} aims to learn a model that can take a pair of graphs and directly ``predict'' a matching between them. Such models extract information for matching from graph features and carry learned knowledge to new graph matching problems. Unlike labels in typical classification problems, the space of possible matchings is combinatorial, which poses difficulties for learning such models. 

Recently there has been remarkable progress in optimizing distributions of discrete structures \citep{maddison2016concrete, paulus2020gradient}. 
This class of methods approximate discrete random variables with continuous ones and then use the reparameterization technique \citep{kingma2013auto,rezende2014stochastic} to optimize the distributions.
In particular, \citet{linderman2018reparameterizing} and \citet{mena2018learning} use this method to learn distributions of permutations and achieve good performances in tasks such as solving jigsaw puzzles. However, learning distributions of matchings is an area left to explore \citep{paulus2020gradient}.

In this work we apply the stochastic softmax trick \citep{paulus2020gradient} to the distribution of matchings and then efficiently learn this distribution through reparameterization. We then use a learning model to parameterize such a distribution to address the graph matching problem, so that the learned model can be applied to any new input graphs. The model is learned to maximize the expected reward under the matching distribution. Comparing to models that directly predict only a single matching, this new model that produces a distribution over matchings is able to explore a wider range of solutions. Furthermore, the stochasticity of the predictive distribution increases the robustness of the learned model, because it is trained to predict a population of good solutions.
 
An optimal matching for a graph pair often cannot be discovered in one shot, and some refinement often improves the quality of the solution. This is particularly true for a learning model because of the generalization error: it is even hard to guarantee that the predicted matching is a local minimum on a new graph pair. Similar observations are also reported in amortized inference \citep{marino2018iterative}.

To address this issue, we also design a learnable architecture that can iteratively refine matchings. In addition to training the model's ability to predict matchings, we also train the model's ability to refine an existing solution for a given graph pair. By maintaining the best matching solution with different refinements, the final prediction will never be worse than the initial prediction.

In addition to our model design, we also have investigated the importance of using dummy nodes \citep{wang2019neural} in graph matching.  Our investigation is motivated by the intended application of matching reactants in molecular reactions. We have the prior knowledge that some nodes cannot be matched. Such nodes can find their place by matching to a dummy node. Our empirical study later indicates the effectiveness of the dummy node on this problem.  

With all these considerations, we develop a unified model, Stochastic Iterative Graph MAtching (SIGMA). We then test this new model on three graph matching tasks. The results indicate that the proposed model produces better matching results than state-of-the-art models. We also do an extensive ablation study of the three elements of the model and show the value of each element in graph matching.

To summarize, our contributions in this work include
\begin{itemize}
\item the design of a stochastic softmax trick for learning a matching distribution; 
\item the proposal of a graph matching model that defines a matching distribution; 
\item the design of iterative refinement for graph matching;
\item the premium performances of the proposed model in three graph matching tasks.   
\end{itemize}

\section{Related Work}
\parhead{Graph Matching.}
Traditionally graph matching has been treated as an optimization problem and addressed by various optimization methods. \citet{livi2013graph,yan2016short,sun2020survey,yan2020learning} have made extensive surveys on this topic. Among these traditional methods, the most related works include sampling methods \citep{lee2010graph,suh2012graph}. However, these sampling methods are usually computationally expensive.

Graph Neural Networks (GNNs) \citep{wu2020comprehensive} were recently used as learning models for graph matching \citep{zanfir2018deep, xu2019cross, wang2019learning, wang2019neural, yu2019learning, fey2020deep, nowak2018revised, rolinek2020deep}. The main idea in this class of work is to use a GNN to encode graph structures into node representation, then nodes are matched based on their vector representation.

\parhead{Distributions of Permutations.}
Recently \citet{mena2018learning} and \citet{paulus2020gradient} have developed new methods of learning distributions of permutations. Sharing the idea of the Gumbel-softmax trick \citep{maddison2016concrete}, these methods devise a continuous sampling procedure that can approximately draw samples of discrete permutation matrices. The sampling procedure allows reparameterization \citep{rezende2014stochastic,kingma2013auto}, which enables efficient optimization of the distribution parameters. This work will develop a new distribution for matching from these distributions.  

\parhead{Iterative Refinement in Learning Models.}
\citet{marino2018iterative,krishnan2018challenges} pointed out that the generalization error of an amortized inference model impedes the inference accuracy on test instances. Then they proposed a learning model that can iteratively refine its solution. This idea is further applied to policy optimization \citep{marino2020iterative}. \citet{chen2020imram} used a similar idea in the task of matching text to images. We will also incorporate the refinement mechanism into the model so it can improve graph matching iteratively.

\section{Background}
Suppose there are a pair of graphs, $G^s=(V^s, E^s, \bX^s)$ and $G^t=(V^t, E^t, \bX^t)$. Here $V^s=\{1, \ldots, n_s\}$ the node set of $G^s$, and $E^s$ is the edge set.
We assume graph nodes have known attributes or feature vectors; let $\bX^s \in \bbR^{n_s \times d}$ denote these node features. Similarly, $V^t$, $E^t$, $\bX^t$, and $n_t$ are, respectively, the node set, the edge set, node features, and the number of nodes of graph $G^t$. Without loss of generality, we always assume $n_s \le n_t$. 

Graph matching identifies a set of node correspondences between $V^s$ and $V^t$. Here we add a ``dummy node'' with id $(n_s + 1)$ to $G^s$ and one with id $(n_t + 1)$ to $G^t$: if some nodes in one graph cannot be matched with a node from the other graph, they will be matched to the dummy node of the other graph. The correspondences between nodes from the two graphs are indicated by a matrix $\bM \in \{0, 1\}^{(n_s+1) \times (n_t+1)}$, 
\begin{align}
\bM &= \left[ \begin{array}{cc}
    \bM^0 & \boldm_t  \\
    \boldm_s^\top & 0
\end{array} \right],  \nonumber\\
 \bone^\top \bM^0 + \boldm_s^\top  &= \bone^\top, \qquad \bM^0 \bone  + \boldm_t  = \bone. 
    \label{eq-dis-matmat}
\end{align}
$M_{i, j} = 1$ indicates node $i$ in $G^s$ is matched to node $j$ in $G^t$. The submatrix $\bM^0$ is the matching between normal nodes in $G_s$ and normal nodes in $G_t$.  
The two equality constraints say every normal node in $G^s$ or $G^t$ must match exactly one node (either a normal node or the dummy node) from the other graph.

Graph matching aims to find a matching $\bM$ that is optimal with respect to an objective $f(\bM; G^s, G^t)$, that is, 
\begin{align}
    \argmax\limits_{\bM} f(\bM; G^s, G^t). 
    \nonumber
\end{align}
In practice, the objective $f(\cdot)$ can be instantiated in many ways. The Quadratic Assignment Problem (QAP) objective \citep{yan2020learning} is: 
\begin{align}
    f_{\mathrm{qap}}(\bM; G^s, G^t) = \sum_{\substack{i, i' \in V^s \\ j, j' \in V^t }} K_{i,j,i',j'}  M^0_{i,j} M^0_{i',j'}.
    \label{eq-signal-qap}
\end{align}
Here a $K_{i,j,i',j'}$ entry can be viewed as the reward of matching a node pair $(i, i')$ in $G^s$ to a node pair $(j, j')$ in $G^t$. Different applications may define different reward functions to emphasize application-specific needs. For example, if each entry $K_{i,j,i',j'}$ is 1 when $(i, i') \in E^s$ and $(j, j') \in E^t$ and 0 otherwise, then the objective count the number of matched edges.

In the supervised learning task, we assume each training pair has a known ground truth matching $\bM^*$. Then the matching $\bM^0$ between normal graph nodes needs to approximate $\bM^*$. \citet{fey2020deep} relax $\bM^0$ to be a continuous variable and compute the negative cross entropy between $\bM^0$ and $\bM^*$. Formally, the objective is defined as
\begin{align}
    f_{\mathrm{sup}}(\bM; \bM^*) = \sum_{(i, j) \in V^s \times V^t} M^*_{i, j} \,\log \left( M^0_{i, j} \right).
    \label{eq-dgmc}
\end{align}

\section{Method}
\begin{figure}[t]
    \centering
    \includegraphics[scale=0.135]{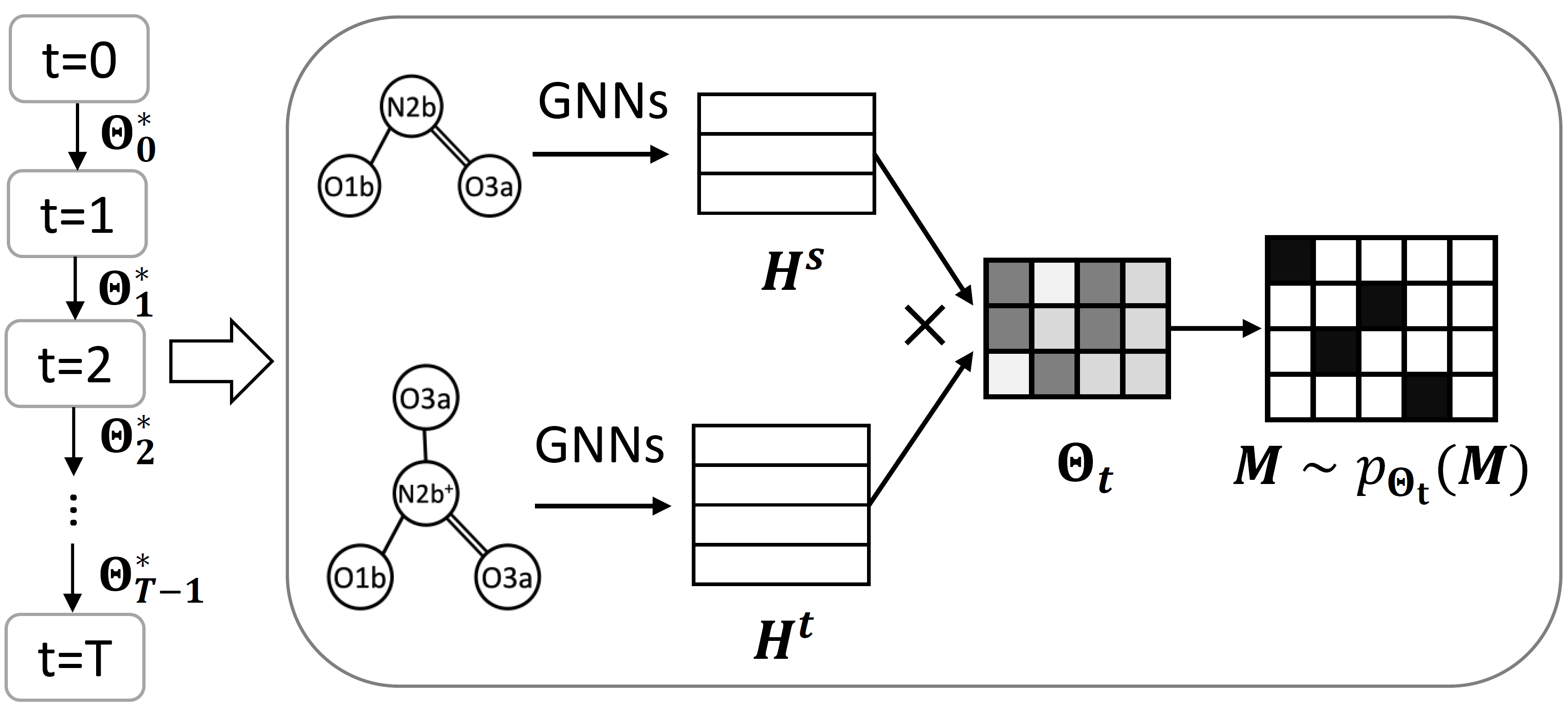}
    \caption{The model architecture of SIGMA. The model predicts $\bTheta^*_0$ that specifies the initial distribution of matchings. Then each refinement iteration continues to improve the distribution in terms of maximizing the expected objective.}
    \label{model-arc}
\end{figure}
The core part of our model is a learnable distribution of the matching matrix $\bM$ between two graphs. We first develop an efficient approach of computing gradients of the distribution parameters. We then develop a learning architecture that can iteratively  refine a predicted matching. An overview of the model is given in Figure \ref{model-arc}.

\subsection{The Matching Distribution} \label{sec-dist}
We develop a model that predicts a distribution of matchings, instead of a single matching, for a training pair of graphs. Suppose the model is parameterized by $\gamma$ and specifies a distribution $p_{\gamma}(\bM | G^s, G^t)$, then we train the model by maximizing the expected matching objective:
\begin{align}
   \max_{\gamma} ~ \calL = \E{p_{\gamma}(\bM | G^s, G^t)}{f(\bM)}.
   \label{eq-obj-ori}
\end{align}
In the testing stage, a predicted matching matrix is the MAP estimate from the distribution. 

We specify the model in two steps: 
\begin{align}
p_{\gamma}(\bM | G^s, G^t)  = p_{\bTheta}(\bM), \quad \bTheta = \mathrm{nn}(G^s, G^t; \gamma).
\label{eq-pgamma}
\end{align}
In the first step, neural network $\mathrm{nn}(G^s, G^t; \gamma)$ computes a parameter $\bTheta$. In the second step, this parameter defines the distribution over matchings $p_\bTheta$. We first consider the form of $p_{\bTheta}$ and then design the network $\mathrm{nn}(\cdot, \cdot; \gamma)$.

We first consider $p_\bTheta$ and the gradient $\partial\calL/\partial \bTheta$ in the optimization \eqref{eq-obj-ori}. Since the integral is often intractable, the gradient with respect to the distribution parameter $\bTheta$ (computed from $\gamma$) is often estimated through Monte Carlo samples. Given that $p_{\bTheta}$ is a discrete probability distribution, an unbiased estimator of $\partial\calL/\partial \bTheta$ is the score function estimator \citep{williams1992simple}, which typically has large variances and often leads to suboptimal results. This work uses the stochastic softmax trick \citep{paulus2020gradient} and derives a low-variance (though biased) estimator of  $\partial\calL/\partial \bTheta$.  

We will find a continuous distribution $\hat{p}_{\bTheta}(\bM)$ to ``imitate'' $p_{\bTheta}(\bM)$: samples from $\hat{p}_{\bTheta}$ are approximately samples from $p_{\bTheta}$.\footnote{Here we use the same notation $\bM$ for random variables in both $p_{\bTheta}$ and $\hat{p}_{\bTheta}$ to reduce the load of notations.} 
Then we can compute the expectation in \eqref{eq-obj-ori} with $\hat{p}_{\bTheta}$ and optimize $\bTheta$ with the reparameterization technique. To achieve this goal, we leverage the Gumbel-Sinkhorn (GS) distribution described below. 

 \paragraph{Discrete distribution over permutations.}
 We can first view the graph matching problem as finding a good permutation.
Let $\bS \in \{0,1\}^{m \times m}$ define a valid binary permutation matrix, which satisfies $\{\bS \bone  = \bone, \bS^\top \bone = \bone \}$. 
To interpret $\bS$ for our problem, we set $m = n_s + n_t$. The rows of $\bS$ are ordered to represent the nodes of $G^s$ followed by $G^t$, while the columns are ordered to represent $G^t$ followed by $G^s$. By selecting this ordering, the top left $n_s \times n_t$ block of $\bS$ determines the matching $\bM_0$ between $G_s$ and $G_t$. The bottom $n_t$ rows of $\bS$ can be viewed as $n_t$ indistinguishable ``dummy nodes'' in $G^s$ so that the $n_t$ nodes in $G^t$ have some chance that they all match to these dummy nodes. Similarly the last $n_s$ columns can be considered as $n_s$ dummy nodes in $G^t$. 

We can define a \emph{distribution} over permutations as
\begin{align}
p_\bPhi(\bS) &\propto \exp\left(\mathrm{trace}\left(\bS^{\top} \bPhi \right)\right). 
\end{align}
where $\bPhi \in \mathbb{R}^{m\times m}$ is a parameter, and the normalization is over the set of valid permutations.

 \paragraph{Relaxation to continuous variables.}
The GS distribution \citep{mena2018learning} is developed to imitate the discrete distributions above but uses continuous random variables that make gradient-based training easier.
The GS distribution $\hat{p}_{\bPhi}(\bS)$ is over the space of doubly stochastic matrices of the same size as $\bS$, and it is reparameterizable with standard i.i.d. Gumbel noise.
We can draw samples from $\hat{p}_{\Phi}(\bS)$ by running a Sinkhorn procedure over a random matrix \citep{mena2018learning}.
The resulting samples can accurately approximate samples from $p_{\bPhi}(\bS)$. 

We define our own distribution $p_{\bTheta}(\bM)$ through $p_{\bPhi}(\bS)$: use $\bTheta$ to decide $\bPhi$, draw samples from $p_{\bPhi}(\bS)$,  and then transform these samples to samples of $p_{\bTheta}(\bM)$. Let's first consider the transformation, then we can decide $\bPhi$ in a meaningful way.
Suppose $\bS$ is a permutation matrix of size $(n_s + n_t)$, then the transformation $\bM = \bB\bS\bC$ gives a sample $\bM$:  
\begin{align}
    \bB = \left[
    \begin{array}{cc}
     \bI_{n_s} &  \\
     & \mathbf{1}_{n_t}^\top
    \end{array}
    \right]&,
    \quad
    \bC = \left[
    \begin{array}{cc}
     \bI_{n_t} &\\
     & \mathbf{1}_{n_s}
    \end{array}
    \right]. \label{eq-transform}
\end{align}
Here $\bI$ is the identity matrix and $\bone$ is a column vector, with sizes indicated by subscripts. The rest of the matrix is all zeros.
After applying the transformation $\bM = \bB \bS \bC$, the top-left $(n_s \times n_t)$ block of $\bM^0$ is the same as in the top-left block of $\bS$, which denotes the correspondence between normal nodes in $G^s$ and $G^t$. The last row and column of $\bM$ simply condense all dummy nodes to one aggregate dummy node in each graph.

With this transformation, the matching distribution $p_{\bTheta}(\bM)$ is automatically defined. 
\begin{align}
  p_{\bTheta}(\bM) = \sum\limits_{\bS: \bM = \bB\bS\bC} p_{\bPhi}(\bS).  \nonumber
\end{align}

With the same linear transformation $\bM = \bB\bS\bC$, we convert a sample from \textit{the continuous distribution} $\hat{p}_{\bPhi}(\bS)$ to a sample from $\hat{p}_{\bTheta}(\bM)$. Then samples of $\hat{p}_{\bTheta}(\bM)$ are approximate samples from $p_{\bTheta}(\bM)$. Because the linear transformation is constant, the approximation error of $\hat{p}_{\bTheta}$ is bounded by the approximation error of $\hat{p}_{\bPhi}$ by a constant factor.  

We now show how to decide the parameter $\bPhi$ from $\bTheta$,  
\begin{align}
    \bPhi =
    \begin{bmatrix}
    \bTheta & \mathbf{0}_{n_s \times n_s}\\
    \mathbf{0}_{n_t \times n_t} & \mathbf{0}_{n_t \times n_s}
    \end{bmatrix}.
    \label{eq-phi}
\end{align}
The matrix $\bTheta$ as a block of $\bPhi$ mainly affects the top-left $(n_s \times n_t)$ block of $\bS$. Every element $\Theta_{i,j}$ indicates the preference of matching $i$ in $G^s$ to $j$ in $G^t$: a large positive value favors the matching while a negative value is against the matching. It is not necessary to differentiate nodes matched to either dummy node, so their corresponding parameters are set to zero.

\parhead{The computation of $\bTheta$.} Finally we design the neural network $\mathrm{nn}(G^s, G^t; \gamma)$ that computes $\bTheta$. A GNN is a powerful model for encoding graph structures into vector forms. GNNs have been previously used for graph matching~\citep{fey2020deep}. We compute the distribution parameter $\bTheta$ via a GNN:
\begin{multline}
\bTheta = \bH^s (\bH^t)^\top, ~~
   \bH^s = \mathrm{GNN}(G^s; \gamma), \\
   ~~ \bH^t = \mathrm{GNN}(G^t; \gamma).
   \label{eq-theta-onestep}
\end{multline}
Here $\bH^s$ and $\bH^t$ are node representations of the graph pair. $\gamma$ denotes all parameters of the $\mathrm{GNN}$. 

Now we have completed the two steps in \eqref{eq-pgamma} and have a learning model $p_{\gamma}(\bM; G^s, G^t)$ for graph matching. To summarize, we use the GNN with weights $\gamma$ to compute $\bTheta$ which in turn determines the parameter $\bPhi$ of the  permutation  distribution. Then, we draw samples from $\hat{p}_{\bPhi}(\bS)$, and apply the transformation $\bM = \bB\bS\bC$ to obtain approximate samples from $p_{\gamma}(\bM; G^s, G^t)$. We can estimate the gradient of an expectation over this distribution with respect to $\bPhi$ using the reparameterization trick. 

\subsection{Iterative Refinement of the Matching}

So far we have a learning model that can predict $\bTheta$ for a pair of graphs.
However, it is difficult to produce an optimal matching for two complex graphs in one step.
A strategy in searching algorithms is to match ``easy'' nodes first and gradually expand the matching. We want our learning model to mimic this strategy and refine $\bTheta$ in multiple steps. 

\parhead{A Refinement Model.} We first design a learning model $\bTheta_1 = \mathrm{nnr}(\bTheta_0; G^s, G^t)$ that can refine a prediction $\bTheta_0$. The goal for $\mathrm{nnr}(\bTheta_0; G^s, G^t)$ is to move $\bTheta_0$ toward  a ``better'' value, that is, increasing the expected objective.   
\begin{align}
    \E{p_{\bTheta_1}}{f(\bM)} \geq \E{p_{\bTheta_0}}{f(\bM)}.
    \label{eq-ite-require}
\end{align}
The model $\mathrm{nnr}(\bTheta; G^s, G^t)$ and the previous network $\mathrm{nn}(G^s, G^t)$ share the same goal: maximizing the expected objective. The difference is that $\mathrm{nnr}(\cdot)$ can get information about the previous matching from  $\bTheta_0$, which helps to
revise prior matchings. 

Ideally the model $\mathrm{nnr}(\cdot)$ should preserve good partial matching and use it to inform further matching of more nodes. Guided by this principle, we give more weights to nodes that are better matched in the previous step. We first compute weight vectors for the two graphs from $\bTheta_0$. Let $\bar{\bM}$ be an average of $\ell$ samples of $p_{\bTheta_0}(\bM)$, and let  
\begin{align}
    \ba_s =  \bar{\bM}_{1:n_s, 1:n_t} \bone,  
    \quad \ba_t =  \bar{\bM}_{1:n_s, 1:n_t}^\top \bone.
    \label{eq-g}
\end{align}
Each entry in vector $\ba_s \in [0, 1]^{n_s}$ indicates the probability that a node in $G^s$ is matched to a normal node in $G^t$. It is similar for the vector $\ba_t \in [0, 1]^{n_t}$. 

Then we use the two vectors to reweight the importance of node features in the GNN computation. 
\begin{align}
\bTheta_1 = \mathrm{nn}(G_u^s, G_u^t), \quad
    G_u^s &= (V^s, E^s, \diag(\ba_s) \bX^s), \nonumber\\ 
    G_u^t &= (V^t, E^t, \diag(\ba_t) \bX^t).
 \label{eq-iterative}
\end{align}
Here the neural network $\mathrm{nn}(\cdot, \cdot)$ is the same one in \eqref{eq-theta-onestep}.
For simplicity, we train one $\mathrm{nn}(\cdot, \cdot)$ model, but we do make sure the model capacity is enough in practice (e.g. try large hidden dimensions and deep models).
Putting \eqref{eq-g} and \eqref{eq-iterative} together, we have the refinement model $\mathrm{nnr}(\bTheta; G^s, G^t)$.

In this design, clustered matched nodes tend to have small changes in their representations so their matching is unlikely to change in the refinement step. Furthermore, these nodes encode the matched structure in their messages to their unmatched neighbors. Note that the message passing in the first layer of the GNN is as if from a weighted graph now. Then the refinement can expand the matching using information from the matched structure. It shares the same principle as searching algorithms \citep{mccreesh2017partitioning}.  

Note that the iterative procedure works on the distribution parameter $\bTheta$ instead of discrete matchings.
The computation $\mathrm{nnr}(\cdot)$ is thus stochastic since $\bar{\bM}$ in \eqref{eq-g} is the average of a small number of samples. The stochasticity allows the model to explore different directions in multiple steps.

\parhead{Multi-step Refinement.} We apply the refinement model to a multi-step procedure. Let $\bTheta^*_0 = \mathrm{nn}(G^s, G^t)$ be the initial  matching distribution. At each step $t = 1, \ldots, T$,  
\begin{align}
\bTheta_{t} &= \mathrm{nnr}(\bTheta^*_{t - 1}; G^s, G^t) \\ 
\bTheta^*_{t} &= \left\{ 
\begin{array}{cc}
\bTheta_t   & \mbox{if } \E{p_{\bTheta_t}}{f(\bM)} \geq \E{p_{\bTheta^*_{t-1}}}{f(\bM)},
\\
\bTheta^*_{t - 1} & \mbox{otherwise.} 
\end{array}
\right.
\end{align}
The two expectations are estimated by samples.
To train the refinement model, we need to track gradients for each $\bTheta^*_{t}$. We do not track gradients with respect to the previous iteration parameter $\bTheta^*_{t-1}$ to stabilize the training procedure.


The refinement procedure will not return a solution worse than the beginning one since the best $\bTheta^*_{t}$ is always kept. If the beginning solution is already very good and hard to improve, then the model will make multiple attempts to improve it. These attempts will not be the same due to the randomness of $\bar{\bM}$. Differing from DGMC's deterministic refinement \citep{fey2020deep}, our refinement is stochastic.
Algorithm \ref{alg-sigma} summarizes our multi-step matching refinement.

\begin{algorithm}[tb]
   \caption{Iterative Refinement}
   \label{alg-sigma}
\begin{algorithmic}[1]
   \STATE {\bfseries Input:} $G^s$, $G^t$, $T$, $f(\cdot)$
  \STATE Initialize $\bTheta_0^*=\mathrm{nn}(G^s, G^t)$
  \FOR{$t=1$ {\bfseries to} $T$}
    \STATE Compute $\bTheta_{t} = \mathrm{nnr}(\bTheta^*_{t - 1}; G^s, G^t)$
    \IF{$\E{p_{\bTheta_t}}{f(\bM)} \geq \E{p_{\bTheta^*_{t-1}}}{f(\bM)}$}
    \STATE $\bTheta^*_{t} = \bTheta_{t}$
    \ELSE
    \STATE $\bTheta^*_{t} = \bTheta^*_{t-1}$
    \ENDIF
  \ENDFOR
  \STATE Return $\left\{\bTheta^*_{0}, \bTheta^*_{1}, \ldots, \bTheta^*_{T}\right\}$
\end{algorithmic}
\end{algorithm}

\subsection{Training and Prediction}
The goal of training is to optimize the parameter $\gamma$, the parameters of the GNN used in both the first-step prediction as well as the refinement procedure.  

This work considers the following matching objective, 
\begin{align}
    f(\bM) = f_{\mathrm{qap}}(\bM) + \lambda f_{\mathrm{sup}}(\bM).
    \label{eq-learning-obj}
\end{align}
Here $f_{\mathrm{qap}}(\bM)$ and $f_{\mathrm{sup}}(\bM)$ are from \eqref{eq-signal-qap} and \eqref{eq-dgmc}.
It covers a wide range of applications. In unsupervised problems, we don't have the second term; in supervised problems, the hyperparameter $\lambda$ balances the importance of the two objectives. We will give more details in the calculation of these objectives in the experiment section. 

After we have included the refinement procedure, the training objective of the entire model for one graph pair $(G_s, G_t)$ becomes 
$\sum_{t=0}^{T} \E{p_{\bTheta_t}}{f(\bM)}$. 
Here each $\bTheta_t$ is computed by the GNN with parameter $\gamma$.
We weigh loss at each step $t$ equally: the first  steps are important because they impact later steps, while the last steps are also important because they are likely to give the final solution.

After training, we need to compute a single matching as the prediction of the model. Our matching model uses the same principle as a typical classification model: using the mode of the predictive distribution as the prediction, though it is harder to find the mode of the matching distribution. The prediction from the distribution $p_{\bTheta^*_T}(\bM)$ is computed from the following problem,
\begin{align}
   \bM &= \argmax\limits_{\bM} ~\mathrm{trace}(\bM^\top \mathrm{pad}(\bTheta^*_{T})) \\ 
   &= \argmax\limits_{\bM = \bB \bS \bC} ~\mathrm{trace}(\bS^\top \bPhi). 
\end{align}
Here $\mathrm{pad}(\bTheta^*_{T})$ adds a row and a column of zeros to $\bTheta^*_{T}$; $\bPhi$ contains $\bTheta^*_T$ as its top-left block as in \eqref{eq-phi}. 
We use Hungarian algorithm to solve the form in the second line. 

Note that when making predictions of the matching, we do not assume any access to ground truth matchings because they will not be available at test time. Thus, our iterative refinement reward function only includes unsupervised terms.

\section{Experiments}
\begin{table*}[ht]
    \centering
    \scalebox{0.88}{\begin{tabular}{ccccccc}
        \toprule
        \textbf{Algorithm} & \textbf{BA(500)$^\S$} & \textbf{BA(500)} & \textbf{BA(1000)} & \textbf{BA(2000)} & \textbf{PPI$^\S$} & \textbf{PPI} \\
        \hline
        MCSPLIT & \textbf{100.0$\pm$0.0} (1) & 43.7$\pm$0.3 (1000) & 16.9$\pm$8.9 (1000) & 33.0$\pm$7.3 (1000) & \textbf{100.0$\pm$0.0} (1) & 60.3 (1000)\\
        S-GWL & 47.5$\pm$0.5 (3) & 32.0$\pm$1.5 (3) & 22.9$\pm$0.3 (14) & 16.2$\pm$0.4 (129) & 83.1 (50) & 81.1 (50) \\
        \hline
        SIGMA & 99.2$\pm$0.2 (10) & \textbf{93.8$\pm$0.3} (11) & \textbf{97.3$\pm$0.2} (33) & \textbf{99.0$\pm$0.1} (179) & 99.2$\pm$0.2 (53) & \textbf{84.7$\pm$0.4} (67)\\
        \bottomrule
    \end{tabular}}
    \caption{Node correctness (\%) and runtime in parenthesis (in seconds). Datasets with $\S$ means 0\% noises; without $\S$ means 5\% noises.}
    \label{exp-mcs}
\end{table*}
We test our model on three tasks: a common graph matching task, a biochemistry application of matching reaction centers among molecular reactants, and a computer vision application of matching keypoints between images.

\subsection{Common Graph Matching} \label{sec-cgm}
\begin{table}[t]
    \centering
    \scalebox{0.88}{\begin{tabular}{ccccc}
    \toprule
    \textbf{Setting} & \textbf{BA(500)} & \textbf{BA(1000)} & \textbf{BA(2000)} & \textbf{PPI}\\
    \hline
    T=0 & 93.5$\pm$0.1 & 97.2$\pm$0.1 & 98.9$\pm$0.1 & 83.1$\pm$0.3\\
    T=3 & 93.7$\pm$0.2 & 97.2$\pm$0.3 & 98.9$\pm$0.1 & 83.8$\pm$0.2\\
    T=4 (default) & 93.8$\pm$0.3 & 97.3$\pm$0.2 & 99.0$\pm$0.1 & 84.7$\pm$0.4\\
    T=9 & 93.8$\pm$0.2 & 97.3$\pm$0.1 & 99.1$\pm$0.1 & 85.0$\pm$0.4\\
    \hline
    Remove S & 93.6$\pm$0.1 & 97.1$\pm$0.2 & 98.9$\pm$0.1 & 84.1$\pm$0.4\\
    \hline
    Remove D & 50.3$\pm$4.1 & 75.0$\pm$5.9 & 80.4$\pm$3.1 & 83.2$\pm$0.2\\
    \bottomrule
    \end{tabular}}
    \caption{Ablation results on common graph matching. Results are reported in node correctness (\%). T: the number of refinement after initial prediction; S: stochastic framework; D: dummy nodes.}
    \label{exp-abalation-general-data}
\end{table}

\parhead{Dataset}
We use two datasets in this task, and follow the experiment setting from \citet{xu2019scalable}. In the first dataset, we use a Barabási-Albert (BA) model to generate graphs of \{500, 1000, 2000\} nodes. To create matching graph pairs, we first sample a source graph from the BA model, then corrupted the source graph by adding $5\%$ noisy edges as a target graph\footnote{We follow the script of S-GWL \citep{xu2019scalable}: \url{https://github.com/HongtengXu/s-gwl}.}. In the second dataset, we start from the Protein-Protein Interaction (PPI) network of yeast (1,004 proteins and 4,920 interactions), and align its $5\%$ noisy version provided in \citet{saraph2014magna}. In both datasets, each node's input feature is assigned according to its node degree. We also include the noise-free versions of the two datasets to match, where the target graph is the same as the source graph.

\parhead{Experiment Setting}
We compare our model with MCSPLIT \citep{mccreesh2017partitioning} and S-GWL \citep{xu2019scalable}. MCSPLIT, which uses branching heuristic to reduce the search space, is a state-of-the-art heuristic method to find an isomorphic subgraph, but it performs poorly when a few ``noise'' edges are added to the graph.
S-GWL, on the other hand, solves the matching problem under the Gromov-Wasserstein discrepancy \citep{chowdhury2019gromov} and has shown robustness to moderate amount of ``noise'' edges added to a graph. For these baselines, we use the authors' implementations with their default hyperparameters. \citet{xu2019scalable} shows that S-GWL has outperformed most heuristic methods on matching noisy graphs.

For our model, we instantiate the GNN as a 5-layer Graph Isomorphism Network (GIN) \citep{xu2018powerful}.
Each layer of GIN has a one-layer MLP with hidden dimension of 256 followed by a $\mathrm{tanh(\cdot)}$ activation.
The model is optimized by an Adam optimizer \citep{kingma2014adam} at a learning rate $10^{-4}$ and trains for 100 epochs. For each dataset, the epoch that produces the best objective is used for testing. We use 10 samples of $\bM$.
$T$ is set to 4 (1 initial prediction followed by 4 iterations of refinement). To evaluate, we report node correctness (NC) as in \citet{xu2019scalable}, which denotes the percentage of nodes that have the same matching as ground truth.

Our model is implemented in PyTorch \citep{paszke2017automatic}. Each model runs on a server with 32 cores and an NVIDIA A100 (40GB) GPU.

\begin{figure*}[t]
    \centering
    \begin{minipage}{0.4\textwidth}
        \centering
    	\includegraphics[width=0.95\linewidth]{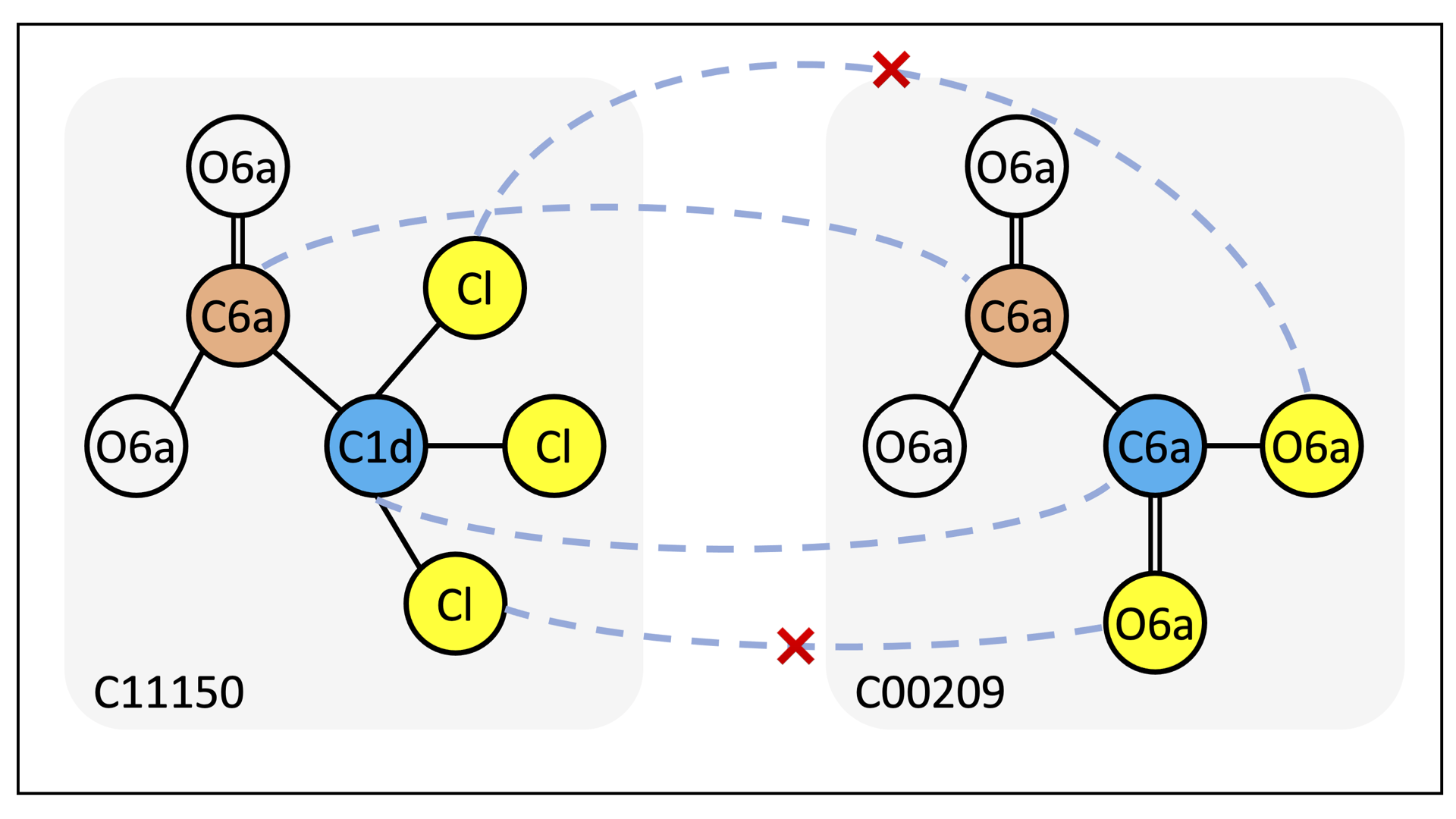}
    \end{minipage}%
    \hfill
    \begin{minipage}{0.425\textwidth}
        \centering
    \includegraphics[width=1.0\textwidth]{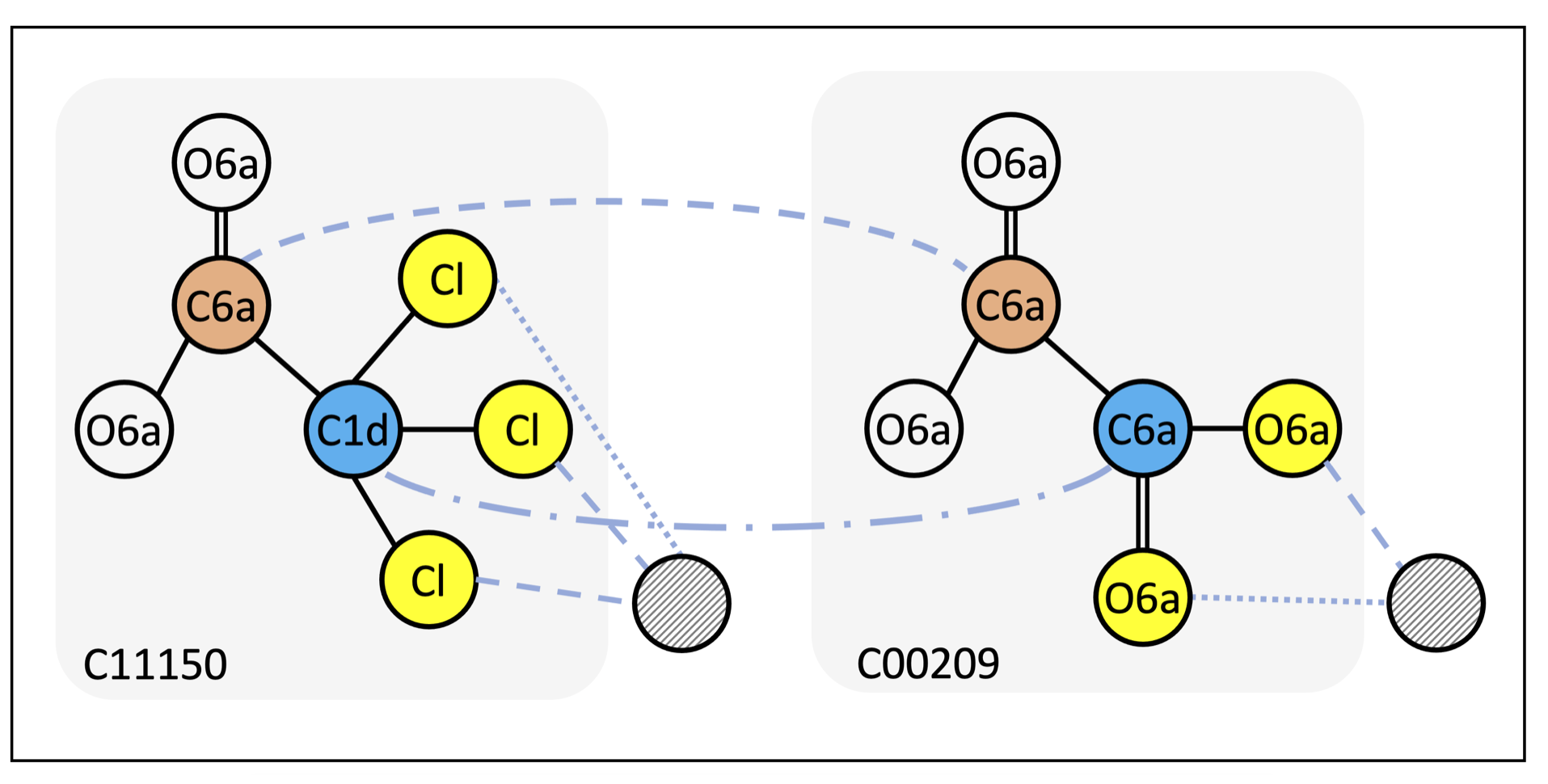}
    \end{minipage}
    \hfill
    \begin{minipage}{0.14\textwidth}
    \centering
    \includegraphics[width=1.0\textwidth]{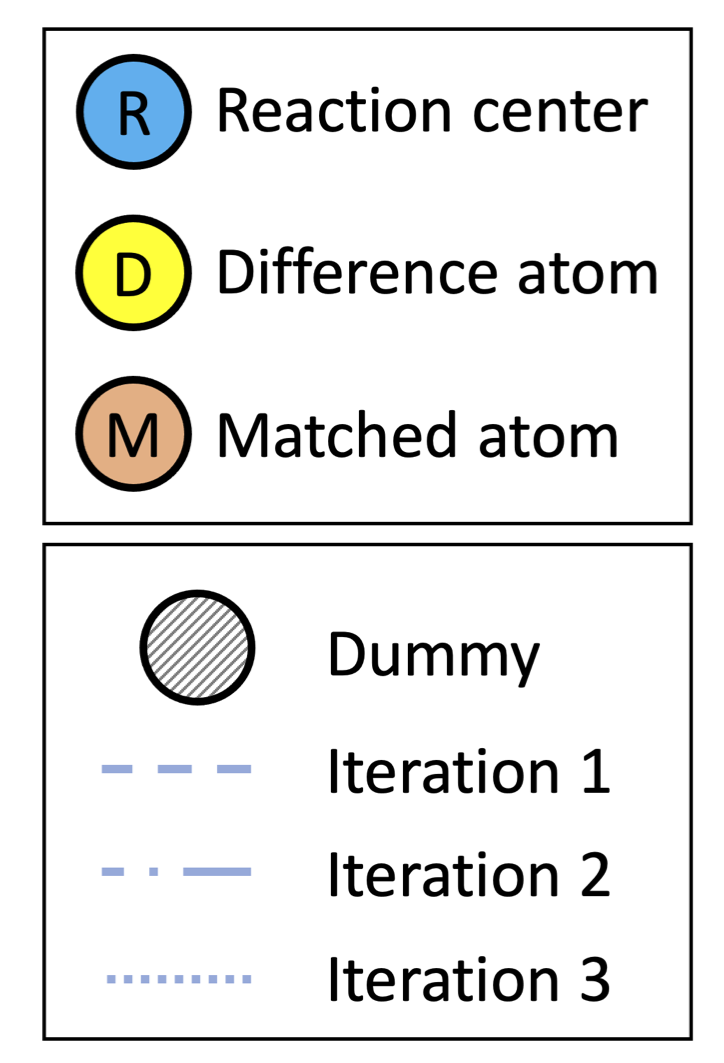}
    \end{minipage}
    \caption{An example of matching RDM between C11150 (Trichloroacetate acid) and C00209 (Oxalate acid). Left: DGMC. Right: SIGMA. In SIGMA, difference atoms (D) are correctly aligned to dummy nodes. SIGMA finds the best match in 3 iterations (1 initial prediction followed by 2 refinements): SIGMA matched \textit{matched atoms (M)} and some \textit{difference atoms (D)} in iteration 1; matched \textit{reaction center (R)} in iteration 2; and matched the rest of \textit{difference atoms (D)} in iteration 3. Dummy nodes are located for the visualization purpose.}
    \label{exp-rdm-example}
\end{figure*}

\parhead{Results}
Results are shown in Table \ref{exp-mcs}. Our model, SIGMA, outperforms baselines in matching noisy graphs. As expected, the heuristic method MCSPLIT fails to cope with noisy graphs. SIGMA outperforms S-GWL on all datasets. We conjecture our improvement stems from learning a GNN that can provide discriminative node embeddings suited for matching. We see comparable matching results between SIGMA and MCSPLIT when matching noise-free graphs. SIGMA does not attain a perfect match; we speculate some nodes lie in symmetric structures in both graphs, and the underlying GNN cannot provide distinguishable embeddings for them. The runtime of SIGMA is competitive to the other two baselines.

We provide an ablation study of the three components: multi-step matching (T), stochastic framework (S), and dummy nodes (D). The ablation study is conducted on noisy graphs to show noticeable performance differences. Results are given in Table \ref{exp-abalation-general-data}. By comparing different $T$, we see more refinement steps yields noticeable improvements on the PPI dataset, though minor improvements on the BA datasets. Using dummy nodes has a clear advantage on this task. When removing the dummy node, the prediction correctness drops up to 43\% on the BA dataset, and 1.5\% on the PPI dataset. Some ``hard'' nodes may have been aligned with dummy nodes and make the matching problem easier. Performance gain using stochasticity is limited in this task.

\begin{table}[t]
    \centering
    \scalebox{0.88}{\begin{tabular}{cccc}
        \toprule
        \textbf{Algorithm} & \textbf{Hard Match} & \textbf{Soft Match} & \textbf{NC}\\
        \hline
        MCSPLIT & 36.0$\pm$1.5 & 54.1$\pm$1.2 & 56.0$\pm$1.2\\
        DGMC & 37.3$\pm$1.5 & 66.0$\pm$1.0 & 66.9$\pm$0.9\\
        \hline
        SIGMA$^U$& 39.5$\pm$1.5 & 63.1$\pm$1.1 & 66.2$\pm$1.0\\
        SIGMA (T=0) & 48.3$\pm$1.5 & 74.8$\pm$0.9 & 76.4$\pm$0.8\\
        SIGMA (\textit{w/o} S) & 50.6$\pm$1.4 & 75.1$\pm$0.9 & 76.0$\pm$0.8\\
        SIGMA (\textit{w/o} D) & 32.3$\pm$1.4 & 59.7$\pm$1.0 & 63.6$\pm$0.9\\
        SIGMA & \textbf{58.0$\pm$1.3} & \textbf{78.2$\pm$0.9} & \textbf{78.3$\pm$0.8}\\
        \bottomrule
    \end{tabular}}
    \caption{Results (in \%) on RDM pattern matching. SIGMA$^U$: Training SIGMA with the QAP objective only (unsupervised).}
    \label{exp-rdm}
\end{table}

\subsection{RDM Pattern Matching in KEGG RPAIR}

\begin{table*}[ht!]
    \centering
    \scalebox{0.62}{
    \begin{tabular}{ccccccccccccccccccccc|c}
        \toprule
        \textbf{Algorithm} & \textbf{Aero} & \textbf{Bike} & \textbf{Bird} & \textbf{Boat} & \textbf{Bottle} & \textbf{Bus} & \textbf{Car} & \textbf{Cat} & \textbf{Chair} & \textbf{Cow} & \textbf{Table} & \textbf{Dog} & \textbf{Horse} & \textbf{M-Bike} & \textbf{Person} & \textbf{Plant} & \textbf{Sheep} & \textbf{Sofa} & \textbf{Train} & \textbf{TV} & \textbf{Mean}\\
        \hline
        GMN & 31.1 & 46.2 & 58.2 & 45.9 & 70.6 & 76.5 & 61.2 & 61.7 & 35.5 & 53.7 & 58.9 & 57.5 & 56.9 & 49.3 & 34.1 & 77.5 & 57.1 & 53.6 & 83.2 & 88.6 & 57.9\\
        PCA-GM & 40.9 & 55.0 & 65.8 & 47.9 & 76.9 & 77.9 & 63.5 & 67.4 & 33.7 & 66.5 & 63.6 & 61.3 & 58.9 & 62.8 & 44.9 & 77.5 & 67.4 & 57.5 & 86.7 & 90.9 & 63.8\\
        CIE & 51.2 & 69.2 & 70.1 & 55.0 & 82.8 & 72.8 & 69.0 & 74.2 & 39.6 & 68.8 & 71.8 & 70.0 & 71.8 & 66.8 & 44.8 & 85.2 & 69.9 & 65.4 & 85.2 & 92.4 & 68.9\\
        DGMC & 47.0 & 65.7 & 56.8 & 67.6 & 86.9 & 87.7 & 85.3 & 72.6 & 42.9 & 69.1 & 84.5 & 63.8 & 78.1 & 55.6 & 58.4 & 98.0 & 68.4 & 92.2 & 94.5 & 85.5 & 73.0\\
        DGMC* & 46.3 & 64.5 & 54.9 & 69.4 & 85.7 & 87.8 & 85.2 & 73.4 & 38.2 & 64.0 & 92.4 & 63.6 & 74.7 & 60.5 & 61.6 & 96.6 & 63.7 & 97.6 & 94.0 & 86.2 & 73.0 \\
        \hline
        SIGMA$^U$ & 26.5 & 43.1 & 40.3 & 66.5 & 85.3 & 86.3 & 73.4 & 43.8 & 29.9 & 40.7 & 94.5 & 33.8 & 61.0 & 39.8 & 52.1 & 95.8 & 40.9 & 95.5 & 92.2 & 84.6 & 61.3\\
        SIGMA (T=0) & 54.0 & 69.9 & 59.9 & 72.3 & 87.4 & 87.9 & 87.0 & 74.0 & 46.2 & 68.5 & 92.4 & 67.9 & 77.3 & 66.9 & 64.4 & 97.2 & 71.6 & 96.7 & 94.7 & 84.6 & 76.0\\
        SIGMA (\textit{w/o} S) & 49.6 & 65.9 & 55.0 & 69.7 & 86.8 & 86.4 & 84.6 & 71.6 & 42.0 & 64.8 & 91.1 & 64.8 & 75.3 & 60.6 & 59.8 & 96.5 & 66.7 & 93.7 & 94.4 & 83.8 & 73.1\\
        SIGMA (\textit{w/o} D) & 56.6 & 71.1 & 59.6 & 71.6 & 87.6 & 83.9 & 89.0 & 74.4 & 46.9 & 71.7 & 86.9 & 69.2 & 80.0 & 69.4 & 64.1 & 96.8 & 71.4 & 96.7 & 94.9 & 87.3 & \textbf{76.5}\\
        SIGMA & 55.1 & 70.6 & 57.8 & 71.3 & 88.0 & 88.6 & 88.2 & 75.5 & 46.8 & 70.9 & 90.4 & 66.5 & 78.0 & 67.5 & 65.0 & 96.7 & 68.5 & 97.9 & 94.3 & 86.1 & 76.2\\
        \bottomrule
    \end{tabular}
    }
    \caption{Hits@1 (\%) on PASCAL VOC with Berkeley annotations. Compared methods are GMN \citep{zanfir2018deep}, PCA-GM \citep{wang2019learning}, CIE \citep{yu2019learning}, and DGMC \citep{fey2020deep}.}
    \label{exp-keypoints-table}
\end{table*}

\begin{figure*}[ht!]
\begin{minipage}{0.72\linewidth}
  \centering
  \begin{minipage}{1.0\linewidth}
    \centering
      {\includegraphics[scale=0.93]{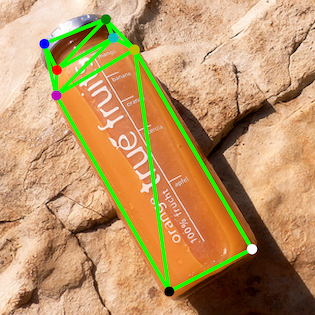}}
      {\includegraphics[scale=0.93]{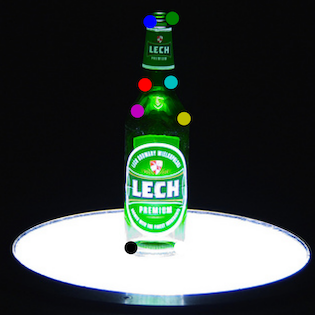}}
      \hspace{1mm}
      {\includegraphics[scale=0.93]{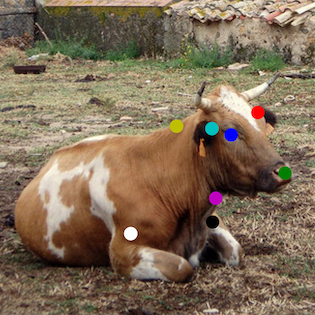}}
      {\includegraphics[scale=0.93]{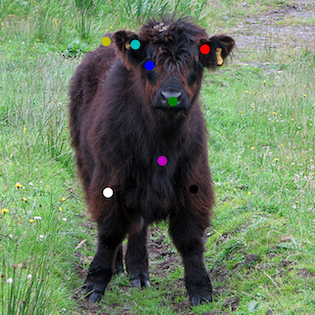}}
      \hspace{1mm}
      {\includegraphics[scale=0.93]{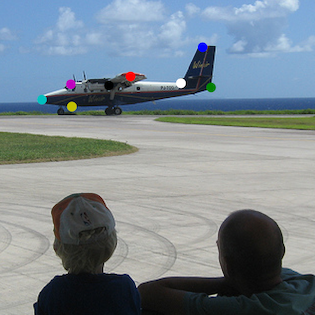}}
      {\includegraphics[scale=0.93]{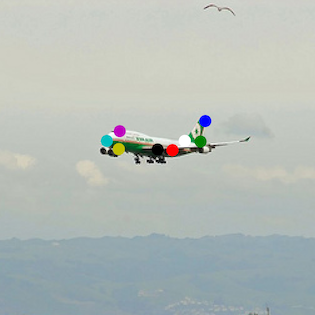}}
      \hspace{1mm}
  \end{minipage}
  \bigskip
  \vspace{-2mm}
  \begin{minipage}{1.0\linewidth}
    \centering
      {\includegraphics[scale=0.93]{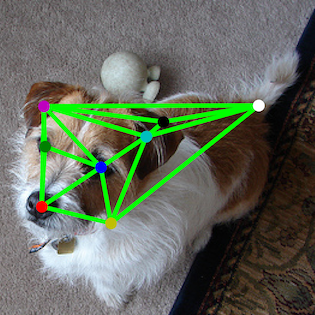}}
      {\includegraphics[scale=0.93]{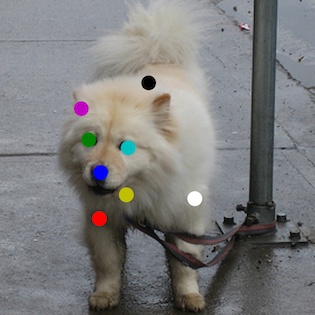}}
      \hspace{1mm}
      {\includegraphics[scale=0.93]{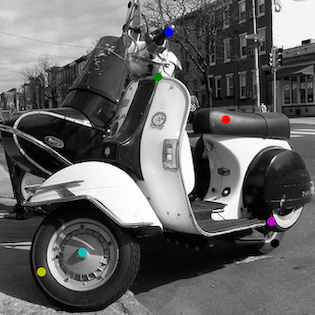}}
      {\includegraphics[scale=0.93]{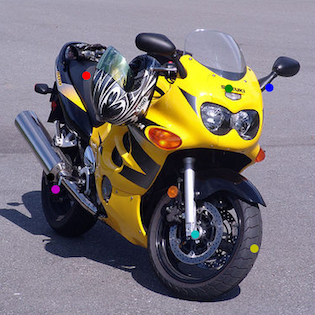}}
      \hspace{1mm}
      {\includegraphics[scale=0.93]{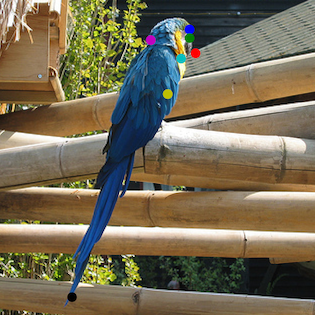}}
      {\includegraphics[scale=0.93]{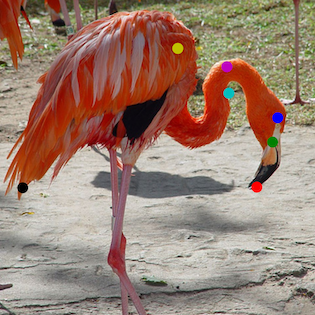}}
      \hspace{1mm}
  \end{minipage}
  \caption{Qualitative examples of SIGMA$^U$ (the first two column) and SIGMA (the last four columns). In each pair of the images, the left shows the source image and the right shows the target. Dots with the same color denote matched nodes. Green lines in the first column are edges of between keypoints. SIGMA correctly identifies keypoints with changed poses.}
  \label{exp-image-show}
 \end{minipage}
 \hfill
 \begin{minipage}{0.25\linewidth}
    \includegraphics[scale=0.48]{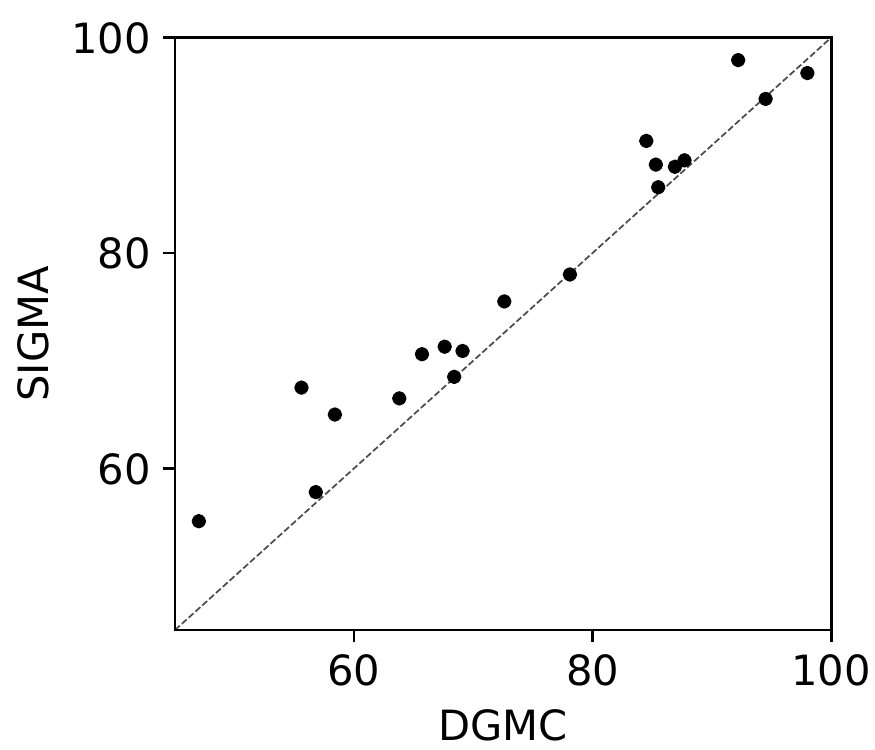}
    \caption{Comparison of Hit@1 \% between SIGMA and DGMC over each category.}
    \label{exp-sigma-dgmc}
 \end{minipage}
\end{figure*}

\parhead{Dataset} 
In the KEGG database\footnote{\url{https://www.genome.jp/kegg/reaction}}, RDM stands for the reaction center atom (R), the difference atoms (D), and the matched atoms (M). RDM patterns record the structural transformation patterns between a reactant pair in an enzymatic reaction \citep{kotera2004computational}. The RPAIR database in KEGG specifies the RDM pattern alignment between two reactant molecules.
The difference atoms are those connected to the reaction center that cannot be matched between the reactants. An example is presented in Figure \ref{exp-rdm-example}. In this task, the aim is to predict the RDM pattern given reactant pairs (and not the overall alignment). 

For each reactant in a reactant pair, a graph is constructed from the corresponding molecular description using  data collected from the KEGG database. Nodes in the graph represent atoms and edges represent bonds. We set node features as the KEGG atom types, which encode for an atomic species (e.g., Carbon, Oxygen, etc) and its neighboring atoms \citep{kotera2004computational}. The edge features are the bond types (single, double, or triple bonds -- three types in total). The ground truth RDM pattern for each reactant pair is a set of node correspondences. These node correspondences are used as the supervised objective during training (the calculation of objective is the ``soft match'' we will define in the experiment setting), and used as the label during testing. We collect 10,366 reactant pairs. On average, each reactant contains 23 nodes and 48 edges. We split the dataset into training, validation, testing at ratio 8:1:1.

\parhead{Experiment Setting}
We compare with MCSPLIT (unsupervised) and Deep Graph Matching Consensus (DGMC, supervised) \citep{fey2020deep}. As edges have discrete features, a 3-layer Relational GCN (RGCN) \citep{schlichtkrull2018modeling} is used as the backbone GNN. We adopt DGMC to use the same GNN structure as ours for a fair comparison. Node features are augmented with a positional feature, which is the eigenvectors of the Laplacian matrix that correspond to the largest 50 eigenvalues (padding zeros otherwise). We found using the positional features improves the matching quality. Separate MLPs are used to embed node's atom feature and position feature. The concatenation of the two resultant embeddings are used as GNN's input. We apply batch normalization between RGCN layers, and use dropout at rate 0.5 and L2 regularizer at weight $10^{-4}$. The model is trained for 30 epochs.
Other settings remain the same as in section \ref{sec-cgm}.

To evaluate, we consider a hard match, a soft match, and the node correctness. For each reactant pair, the hard match denotes whether the prediction fully matches the RDM pattern; it takes value 1 or 0. The soft match is a relaxed version of the hard match; it shows the fraction of node pairs matched in RDM and takes a value between 0 and 1. During training, the soft match is the supervised objective.

\parhead{Results} Results are given in Table \ref{exp-rdm}. SIGMA attains the top performance across all metrics. We also run SIGMA in an unsupervised setting (denoted as SIGMA$^U$), where SIGMA is trained with the QAP objective only. As expected, the performance degenerates. This result emphasizes the demand of designing a learning model that supports various objective function. Ablation results show each of our three components plays a vital role.

In Figure \ref{exp-keypoints-table}, we illustrate a comparison between DGMC and SIGMA. SIGMA correctly matches the RDM pattern, where difference atoms are aligned to dummy nodes. The result further shows the effectiveness of our multi-step refinement: the refinement process matches the three different RDM regions in a progressive way.

\subsection{Image Keypoints Matching}

\parhead{Dataset} In this task, we match keypoints on PASCAL VOC \citep{everingham10} with Berkeley keypoint annotations \citep{bourdev2009poselets}. The dataset is difficult, because it contains images of various scales, poses and illuminations \citep{wang2019learning}. In total, the dataset contains 20 classes of objects. On average, each class contains 348 training graphs and 84 test graphs, where each graph contains 1 to 12 keypoints and 1 to 27 edges. We  follow the experimental settings as Deep Graph Matching Consensus (DGMC) \citep{fey2020deep}. The original dataset is filtered to 6,953 images for training and 1,671 images for testing. Difficult, occluded, and truncated objects are excluded, and each image has at least one keypoint. For keypoints features, we use a concatenation of the output of \texttt{relu4\_2} and \texttt{relu5\_1} from a pre-trained VGG16 \citep{simonyan2014very} on ImageNet \citep{deng2009imagenet}. Edge features are normalized 2D Cartesian coordinates (the anisotropic setting from DGMC).

\parhead{Experiment Setting} We compare with four state-of-the-art baselines: GMN \citep{zanfir2018deep}, PCA-GM \citep{wang2019learning}, CIE \citep{yu2019learning}, and DGMC \citep{fey2020deep}.
Following DGMC, we set our backbone GNN as a SplineCNN \citep{fey2018splinecnn}. Most of the hyperparameters are kept the same as DGMC. The kernel size is 5 in each dimension. We stack 2 convolutional layers, followed by a dropout layer with probability 0.5, which in turn is  followed by a linear layer that outputs node embeddings $\bH$. Unlike DGMC, we found a hidden dimension of 512 and a LeakyReLU activation with a negative slope 0.1 work well for our model (DGMC uses 256 hidden layers and a ReLU activation). For a fair comparison, we also run DGMC with this setting and name it as DGMC$^*$. Following DGMC, the supervised objective is defined as in \eqref{eq-dgmc}. We report Hit@1 to evaluate the performance. Hit@1 shows the percentage of correct matched instances over the whole.

\parhead{Results} Results are given in Table \ref{exp-keypoints-table}. SIGMA significantly outperforms GMN and PCA-GM by over 10\% of average Hit@1 score, and a 7.3\% improvement upon CIE. We also observe a 3.2\% of Hit@1 improvement over the DGMC. In Figure \ref{exp-sigma-dgmc}, we further confirm SIGMA produces better matching over most of the categories. In this dataset, the stochastic framework brings the most performance gain. Removing the stochasticity, which means optimizing (through relaxation) a single matching $\bM$ in \eqref{eq-dis-matmat}, drops the Hit@1 score by 3.1\%. On average, the effectiveness of dummy nodes and multi-step matching is minor on this task.

Our unsupervised setting, SIGMA$^U$, shows a 3.4\% improvement over GMN on the average score  and is on par performance with PCA-GM. Note that both GMN and PCA-GM are supervised. The QAP objective determines the predictive capacity of SIGMA$^U$. Once the QAP objective can recognize the input graph's topology, SIGMA$^U$ has the potential to perform matching well. In the first two columns of Figure \ref{exp-image-show}, we see SIGMA$^U$ successfully matches the bottle image pair but partially matches the dog image pair. We guess that the bottle image pair presents a more recognizable graph structure to the QAP objective (edges in green lines) than the dog image pair. The last four columns in Figure \ref{exp-image-show} visualize four correctly matched samples from SIGMA. SIGMA recovers node correspondences under various pose changes.

\begin{table}[t]
    \centering
    \scalebox{0.8}{\begin{tabular}{ccc}
    \toprule
    \textbf{Algorithm} & \textbf{PASCAL VOC} & \textbf{SPair-71k}\\
    \hline
    BB-GM & 80.1$\pm$0.6 & 78.9$\pm$0.4\\
    SIGMA & \textbf{81.2$\pm$0.2} & \textbf{79.8$\pm$0.2}\\
    \bottomrule
    \end{tabular}}
    \caption{Mean Hits@1 (\%) on  PASCAL VOC and SPair-71k. Compared method is BB-GM \citep{rolinek2020deep}. SIGMA results are reported following BB-GM's experiment setting.}
    \label{exp-comp-bbgm}
\end{table}

Lastly, we compare SIGMA with recent method BB-GM \citep{rolinek2020deep} using BB-GM's representation learning method. Note that BB-GM focuses on learning representations and uses a match solver as a blackbox. We follow BB-GM's experiment setting, and the main differences from the previous experiment include: 1) fine-tuning VGG16's weights and 2) computing node affinities $\bTheta$ from a weighted inner product (the weights of the inner product are from the final VGG16 layer). 
Then we evaluate both models on PASCAL VOC and SPair-71 \citep{min2019spair}. SPair-71k is similar to PASCAL VOC, but contains higher quality images and richer keypoints annotations. Table \ref{exp-comp-bbgm} shows that SIGMA outperforms BB-GM in terms of mean Hits@1.

\section{Conclusion}
We have introduced a new learning model, SIGMA, that addresses graph matching problems. We presented two innovations in the design of this new model. First, the model learns a distribution of matchings, instead of a single matching, between a pair of graphs. Second, the model learns to refine matchings through attending to matched nodes. SIGMA consistently shows  better performance than other methods
in terms of the matching quality, and at the same time, retains a comparable running speed as the baselines. 

SIGMA opens many possible directions. Since matchings sampled from SIGMA are still like discrete variables, high-level graph structure can still be well defined on these samples. Therefore, SIGMA can be applied to match high-order graph structures such as paths and hyper-edges.  With slight modification, SIGMA can be applied to matching problems beyond graph matching, such as matching tabular data to knowledge graphs.   
We hope that this work not only brings a new tool for graph matching but also inspires further research in this direction.

\section*{Acknowledgements}
We thank all reviewers for their insightful comments. This research was supported by NSF 1908617 and NSF 1909536. Soha Hassoun's research
is also supported by the NIGMS of the National Institutes of Health, Award
R01GM132391. The content is solely the responsibility of the authors and
does not necessarily represent the official views of the NIH.

\bibliographystyle{apalike}
\bibliography{main}

\end{document}